\documentclass[letterpaper, 10 pt, conference]{ieeeconf}
\IEEEoverridecommandlockouts    
\usepackage{graphics}           
\usepackage{times}              
\usepackage{amsmath}            
\usepackage{amssymb}     
\usepackage{mathrsfs}
\usepackage{graphicx}
\usepackage{algorithm}
\usepackage{url} 
\usepackage[noend]{algpseudocode}
\usepackage{booktabs}
\usepackage{color}
\usepackage[nocompress]{cite} 
\usepackage{multirow} 
\usepackage{censor}
\definecolor{instructioncolor}{rgb}{.5,.5,.5}

\usepackage[font=small]{caption}

\def\eqref#1{Eq.~(\ref{#1})}

\makeatletter
\usepackage{xspace}
\DeclareRobustCommand\onedot{\futurelet\@let@token\@onedot}
\def\@onedot{\ifx\@let@token.\else.\null\fi\xspace}

\def\etal{{et al}\onedot}
\makeatother

\def\etalcite#1{\etal~\cite{#1}}

\usepackage{array}
\newcolumntype{L}[1]{>{\raggedright\let\newline\\\arraybackslash\hspace{0pt}}m{#1}}
\newcolumntype{C}[1]{>{\centering\let\newline\\\arraybackslash\hspace{0pt}}m{#1}}
\newcolumntype{R}[1]{>{\raggedleft\let\newline\\\arraybackslash\hspace{0pt}}m{#1}}

\usepackage{flushend} 

\title{\LARGE \bf Keypoint Semantic Integration for Improved Feature Matching\\ in Outdoor Agricultural Environments}

\author{Rajitha de Silva$^{1}$, Jonathan Cox$^{1}$, Marija Popovi\'{c}$^{2}$, Cesar Cadena$^{3}$, Cyrill Stachniss$^{4}$ and Riccardo Polvara$^{1}$
  \thanks{$^{1}$Rajitha de Silva, Jonathan Cox and Riccardo Polvara are with Lincoln Centre for Autonomous Systems (L-CAS), University of Lincoln, UK. $^{2}$Marija Popovi\'{c} is with MAVLab, TU Delft, Netherlands. $^{3}$Cesar Cadena is with Robotics Systems Lab, ETH Zurich, Switzerland. $^{4}$C. Stachniss is with the University of Bonn, Germany. (correspondence author: Rajitha de Silva {\tt\small $^{1}$odesilva@lincoln.ac.uk})}%
  \thanks{This work was supported by Engineering and Physical Sciences Research Council (EPSRC), UK Project: ``GAIA: Ground-Aerial maps Integration for increased Autonomy outdoors" (EPSRC Reference: EP/Y003438/1).}
}
\begin{document}
\maketitle
\thispagestyle{empty}
\pagestyle{empty}

\begin{abstract}
  %


  
Robust robot navigation in outdoor environments requires accurate perception systems capable of handling visual challenges such as repetitive structures and changing appearances. Visual feature matching is crucial to vision-based pipelines but remains particularly challenging in natural outdoor settings due to perceptual aliasing.
We address this issue in vineyards, where repetitive vine trunks and other natural elements generate ambiguous descriptors that hinder reliable feature matching. We hypothesise that semantic information tied to keypoint positions can alleviate perceptual aliasing by enhancing keypoint descriptor distinctiveness. To this end, we introduce a keypoint semantic integration technique that improves the descriptors in semantically meaningful regions within the image, enabling more accurate differentiation even among visually similar local features.
We validate this approach in two vineyard perception tasks: (i)~relative pose estimation and (ii)~visual localisation. Across all tested keypoint types and descriptors, our method improves matching accuracy by 12.6\%, demonstrating its effectiveness over multiple months in challenging vineyard conditions.
\end{abstract}

\section{Introduction}
\label{sec:intro}

Modern agricultural systems require robust robotic solutions to automate labour-intensive tasks and enhance operational efficiency. Among these, vision-based perception gaining traction due to its affordability. For autonomous mobile robots operating in outdoor agricultural environments, perception systems must accurately detect and track visual features for tasks such as relative pose estimation and visual localisation. However, vineyards present repetitive structures—vine trunks, poles, water pipes—that complicate feature correspondence between consecutive frames. Both traditional and learned keypoint descriptors focus on local regions, lacking broader semantic context, which can lead to false-positive matches for spatially distant but visually similar keypoints. This challenge, known as perceptual aliasing, remains a major hurdle towards dependable vision-based navigation in outdoor settings. 

\begin{figure}[t]
  \centering
  \includegraphics[width=0.99\linewidth]{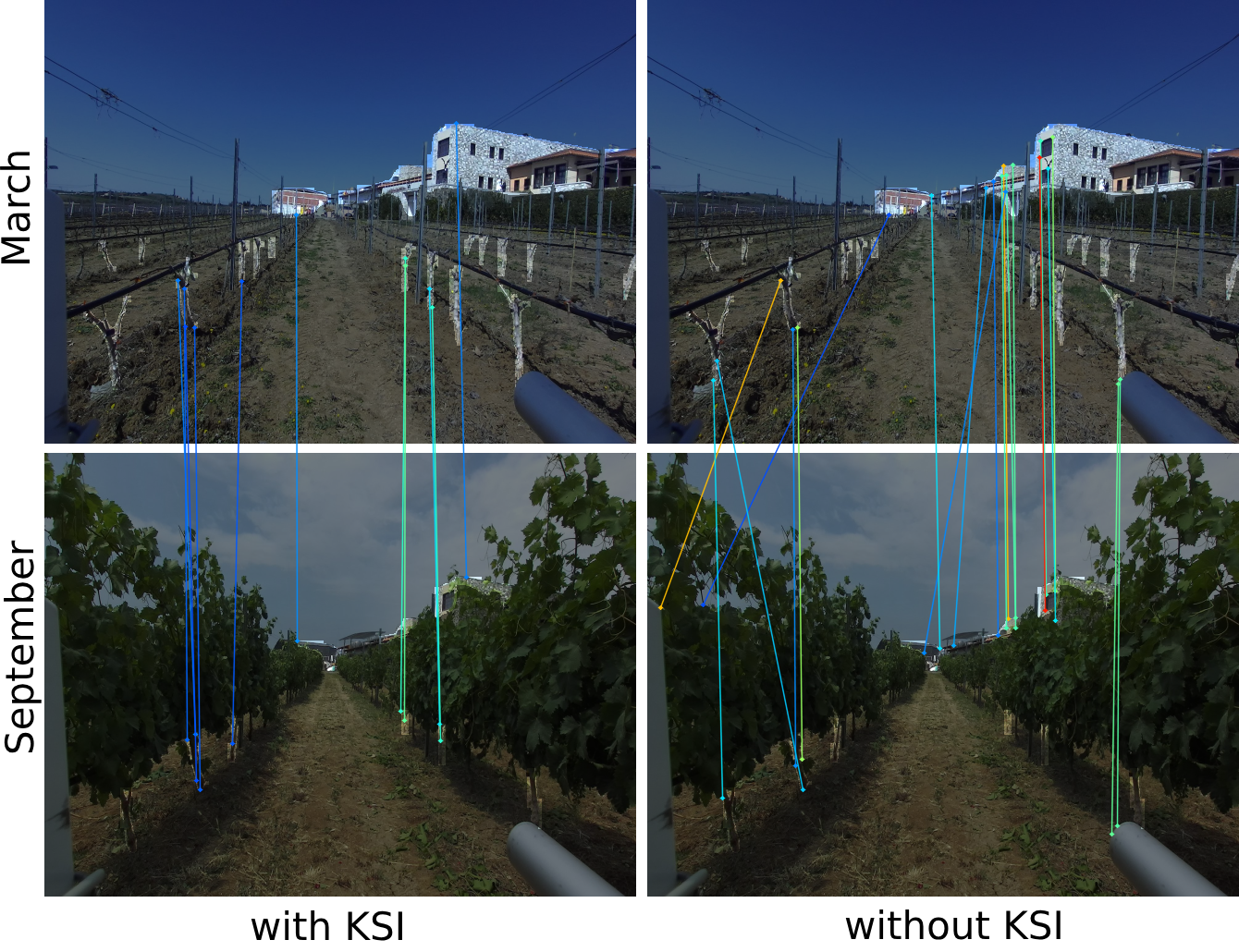}
  \caption{KSI enhances the keypoint descriptors with corresponding semantic instances to minimise aliasing during keypoint matching in vineyards. Keypoint matches on semantic masks are visualised here across March and June.}
  \label{fig:motivation}
\end{figure}

In this paper, we address this challenge by exploiting the semantic and morphological properties of repetitive regions to reduce descriptor ambiguity and improve matching accuracy. Our approach enriches keypoint descriptors with semantic context to increase their distinctiveness, focusing on vineyard environments where repetitive structures are widely recognised yet rarely supplemented with semantic cues for descriptor differentiation~\cite {9812419}. Vine trunks, for instance, appear visually similar, producing nearly identical descriptors; however, morphological variations (Fig.~\ref{fig:motivation}) can be encoded with semantic embeddings to refine keypoint representations. This approach supports accurate keypoints matching across multiple months, even under significant seasonal changes.

Typical feature matching pipelines rely on heuristic filtering~\cite{9812419} or incorporate context at the matching stage~\cite{Sarlin_2020_CVPR}, but they still struggle with uniform descriptors and repetitive spatial layouts~\cite{chebrolu2018ral, wang2024benchmarking}. We propose keypoint semantic integration (KSI), a framework for enhancing the keypoint descriptors with semantic mask instances on which they are located. An autoencoder~\cite{8616075} compresses the semantic mask into a feature vector, generating a semantic embedding that enhances descriptors' distinctiveness. Consequently, keypoints on semantically meaningful regions gain enriched descriptors, while those in the background retain their original descriptors.



Our contributions are are: (i) a semantic integration method, KSI\footnote[1]{To support reproducibility and future research, our implementation and dataset will be available at: [Link will be available at acceptance]}, that mitigates perceptual aliasing in visually repetitive agricultural environments; (ii) an extensive ablation and evaluation study demonstrating the method’s robustness under seasonal changes in vineyards; and (iii) the Semantic Bacchus Long Term (SemanticBLT) dataset with panoptic segmentation annotations in vineyards.


\section{Related Work}
\label{sec:related}



\textbf{Visual feature extraction} and matching are essential first steps in a traditional visual simultaneous localisation and mapping (SLAM) pipeline. Classic feature extractors, such as Scale Invariant Feature Transform (SIFT)~\cite{790410} and Oriented FAST and Rotated BRIEF (ORB)~\cite{6126544}, remain popular for their rotation and scale invariance and computational efficiency, while deep learning-based feature extractors like SuperPoint~\cite{DeTone_2018_CVPR_Workshops}, D2-Net~\cite{d2net}, and Reliable and Repeatable Detector Descriptor (R2D2)~\cite{r2d2}) prioritise robustness, repeatability, and reliability. However, neither traditional nor deep learning–based descriptors inherently capture the semantic context of a keypoint, instead relying on local neighbourhood information of the keypoint. Variants such as Semantic SuperPoint~\cite{9996027} combine semantic information with descriptor learning, but must be retrained for each new semantic class, hindering generalisation capabilities. Other approaches integrate semantic cues at the matching stage or via heuristic filtering~\cite{9812419}, adapt descriptors online based on the current environment~\cite{guadagnino2022ral}, or exploit sequence information to address repetitive scenes in agricultural environments~\cite{lobefaro2023iros}.

\textbf{Keypoint matching} in agricultural environments poses significant challenges due to the scarcity of unique landmarks, repetitive crop appearances, lightning changes, and uneven terrain~\cite{chebrolu2018ral, wang2024benchmarking}. Seasonal changes and foliage growth~\cite{hroob2021benchmark} exacerbate these issues, leading to perceptual aliasing and matching errors. Classic matching algorithms, such as nearest neighbour search, struggle to address these issues effectively. Ways to address this include exploiting geometric structures of keypoints~\cite{chebrolu2018ral} or exploiting background knowledge of the agricultural domain~\cite{lobefaro2023iros,lobefaro2024iros}.  Learning-based matches e.g., SuperGlue~\cite{Sarlin_2020_CVPR}, LightGlue~\cite{Lindenberger_2023_ICCV} harness graph neural networks (GNNs) or transformers to incorporate local context for more robust matching. Tanco \etalcite{tanco2023learning} highlight the limitations of traditional detection algorithms through an agriculture-specific feature descriptor and matching method. However, vineyard-specific solutions remain limited \cite{slamagreview}. Papadimitriou \etal~\cite{9812419} address this issue by restricting keypoint extraction to semantically significant regions, such as vine trunks, which retain a relatively consistent appearance over time and remain visible despite foliage growth. Their approach, however, simply masks these regions rather than integrating explicit semantic context into the descriptors.

\textbf{The use of semantic cues} in keypoint detection and matching has been explored in a variety of contexts, including urban environments~\cite{kreiss2021openpifpaf}, remote sensing~\cite{cao2024tsk}, robotic manipulation~\cite{luo2022skp}. and agriculture~\cite{tanco2023learning}, where semantic information is learned from different object classes. Yet most existing semantic keypoint extractors rely on a multi-task backbone~\cite{xue2023sfd2} with separate heads for semantic labels and keypoint descriptors, limiting them to the classes encountered during training. Tanco \etalcite{tanco2023learning} observe that, while their approach is trained with a semantic decoding head, its performance in inferring semantic labels at test time remains suboptimal — suggesting opportunities to better understand the impact of this multi-task setup on keypoint descriptors compared to standard descriptor-based methods. These limitations accentuate the need for a modular approach that extends easily to new semantic classes, improving generalisation in specialised contexts such as agriculture.

\begin{figure*}[t]
  \centering
  \includegraphics[width=0.99\linewidth]{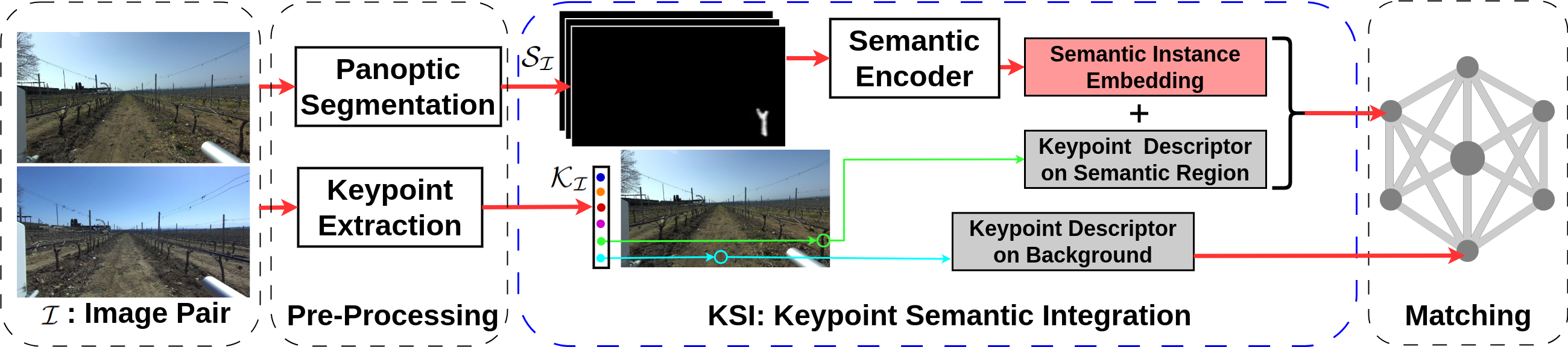}
  \caption{Overview of proposed KSI framework: Input semantic instances $\mathcal{S_I}$ and the set of keypoints $\mathcal{K_I}$ are extracted from each image in the input image pair $\mathcal{I}$. The descriptors of keypoints on a semantic instance are combined with the corresponding semantic embedding to generate semantically enriched keypoint descriptors while the background descriptors are left unaltered. The semantically enriched descriptors and the background descriptors are matched together in a shared matcher. 
  }
  \label{fig:overall}
\end{figure*}

\textbf{Datasets} that represent vineyard-specific semantic classes are crucial for integrating semantic information into keypoint descriptors. The ARD-VO dataset \cite{crocetti2023ard} contains vineyard and olive grove images but offers no semantic segmentation and covers only August to October, when the foliage is fully developed. In contrast, the Bacchus Long Term dataset~\cite{blt} spans from March to September, capturing conditions both before and after major foliage growth. Its initial release (BLT) lacked annotations, so we extend it to Semantic-BLT, adding images and matching semantic masks for training segmentation models. Large, multi-month datasets are critical for robust performance in low-light and dense vegetation \cite{schmidt2024rover}; Semantic-BLT fills this gap in vineyard environments.

Unlike existing methods, our approach provides a modular way to embed semantic cues in keypoint descriptors, supporting both classical and learned descriptors as well as diverse matching algorithms. We use a stand-alone panoptic segmentation model to produce instance masks, which are then converted into semantic embeddings. Unlike systems requiring retraining of keypoint extractors for specific semantic classes, ours requires only semantic segmentation, thus offering flexibility while boosting descriptor distinctiveness in semantically crucial vineyard regions. These enriched descriptors, in turn, enhance both pose estimation and visual localisation in vineyard environments.

\section{Our Approach\\to Keypoint Semantic Integration}
\label{sec:main}
We generate semantically enhanced descriptors by adding semantic embeddings, generated with an autoencoder trained on instance masks, to the original descriptor. In our KSI architecture (Fig.~\ref{fig:overall}), we apply these enhancements only to keypoints falling on semantic instance masks, while background keypoints retain their original descriptors. Although developed for vineyard environments, this method could generalise to other visually repetitive domains, such as orchards, forests, or scree slopes.

\subsection{SemanticBLT Dataset}
\label{ssec:sblt}
Vineyards undergo noticeable visual change, requiring a multi-month dataset for segmentation models to handle diverse foliage, lighting, and environmental conditions. To this end, the SemanticBLT dataset labels six classes (buildings, pipes, poles, robots, trunks, and vehicles) across 1035 images captured from March to September in a single vineyard.

The data collection procedure employed the Thorvald robotic platform with a forward-facing ZED2 camera among other sensors.  Whereas the BLT dataset~\cite{blt} focused on lidar and unannotated image data, SemanticBLT offers semantically segmented imagery suited for vision-based tasks in repetitive agricultural environments.

\subsection{Keypoint Extraction}
\label{ssec:kx}
The KSI architecture enriches existing keypoints by embedding semantic instance mask information in a modular manner compatible with various keypoint extractors. We use SuperPoint descriptors by default, but R2D2~\cite{r2d2}, ORB~\cite{6126544}, and SIFT~\cite{790410} can also be integrated into SuperGlue-based feature matching~\cite{Sarlin_2020_CVPR}. In contrast,  LightGlue matcher required pre-trained weights for each descriptor type, limiting its adaptability. From an image pair $\mathcal{I}$, we extract the keypoints ${K_I = \{(pi, d_i)}\}$, where $p_i \in \mathbb{R}^2$ represents the spatial coordinates and $d_i \in \mathbb{R}^D$ the corresponding descriptors.


\subsection{Panoptic Segmentation in Vineyards}
\label{ssec:ps}
\begin{figure*}[t]
  \centering
  \includegraphics[width=0.99\linewidth]{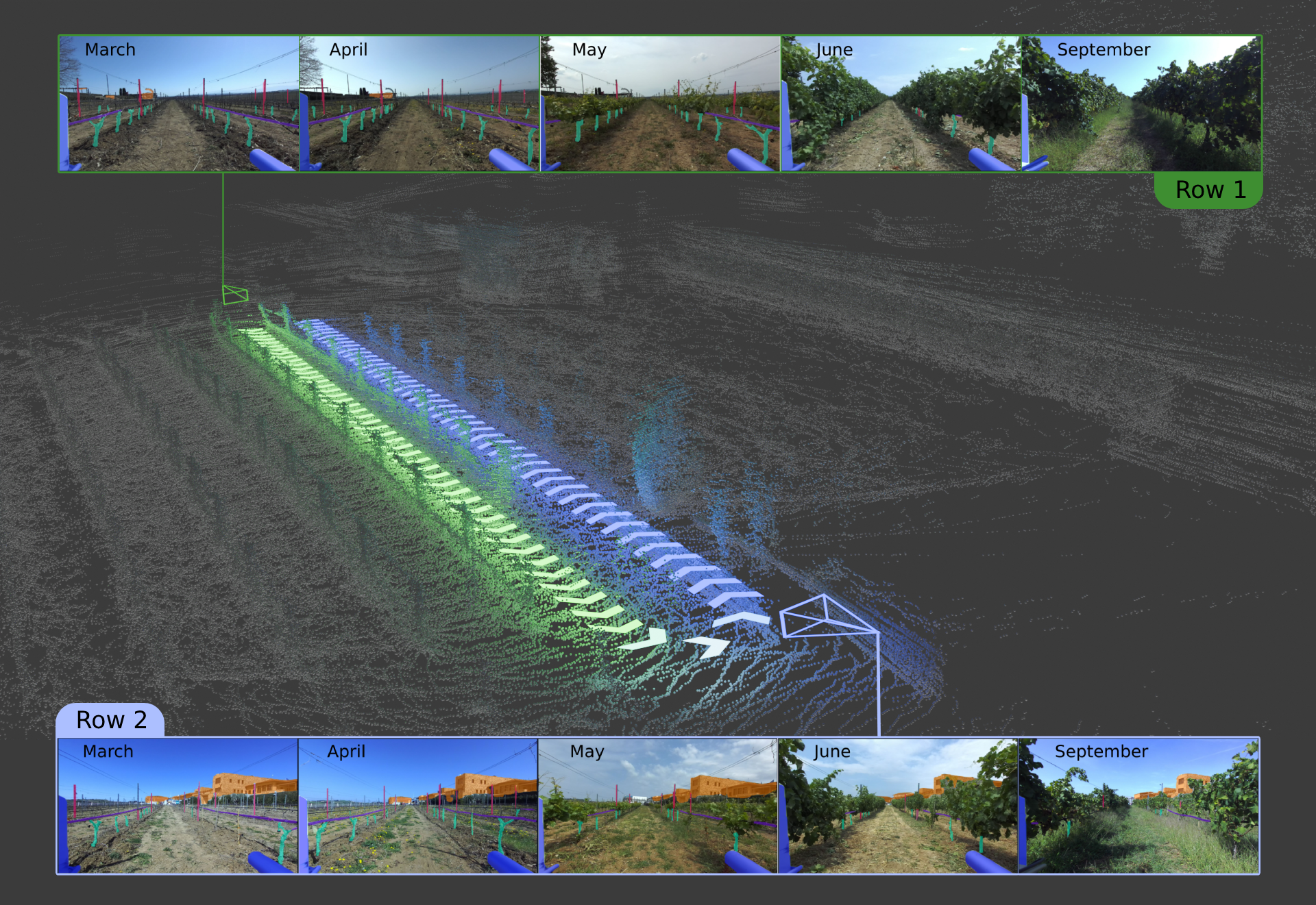}
  \caption{Visualization of the vineyard loop path from which the data for experiments described in Section \ref{sec:exp} was colllected. The robot traverses row 1 in the forward direction, then returns along row 2, completing a closed-loop trajectory. Despite foliage occlusions, trunk and building classes remain visible across all months, rendering them ideal for enhancing keypoint descriptors. Buildings: Orange, Pipes: Purple, Poles: Red, Robot: Blue, Trunks: Cyan.}
  \label{fig:vin}
\end{figure*}

To capture raw semantic and spatial information, we apply a YOLOv9~\cite{yolov9} instance segmentation model trained on SemanticBLT to generate masks for six vineyard-relevant classes (buildings, pipes, poles, robots, trunks, vehicles). The panoptic segmentation component is modular, allowing the YOLOv9 model to be replaced with any other segmentation model without affecting the overall implementation. For KSI, we focus on trunks and buildings, whose distinct shapes and textures support accurate matching. In contrast, pipes and poles are excluded due to uniform shapes that often produce mismatches, while robots and vehicles are ignored since they are not guaranteed to remain static.
Figure~\ref{fig:vin} illustrates the variation of all these semantic classes across multiple months. 
This panoptic segmentation step outputs a set of instance masks $\mathcal{S_I}$ from the image pair $\mathcal{I}$.

\subsection{Semantic Instance Encoding}
\label{ssec:se}

We partition keypoints $\mathcal{K_I}$ into two subsets: those on semantic masks $\mathcal{K_S}$ and those in the background $\mathcal{K_B}$. Each keypoint in $\mathcal{K_S}$ is assigned a semantic embedding via a convolutional autoencoder $[f_{enc},f_{dec}]$ trained on SemanticBLT instance masks. The encoder $f_{enc}$ maps an instance masks $s_i$ to a feature vector capturing morphological cues (shape, size, position). We add this embedding to the original descriptor, then L2-normalise ($N_{L2}(\cdot)$) to yield enhanced descriptors $\mathcal{K'_S}$, preserving the original descriptor dimensions $D \times 1$:


\begin{equation}
\label{eq:sem}
\begin{aligned}
\mathcal{K'_S} &= \left\{ (\mathbf{p}_i, \mathbf{d}'_i) \right\}\\ 
 &=  \left\{(\mathbf{p}_i, N_{L2}\left(\mathbf{d}_i + f_{enc}(\mathbf{s}_i)\right), \, \forall (\mathbf{p}_i, \mathbf{d}_i) \in \mathcal{K_S}) \right\}.
\end{aligned}
\end{equation}

\subsection{Descriptor Matching}
\label{ssec:dm}

The KSI framework, illustrated in Fig.~\ref{fig:overall}, produces two sets of keypoints -- $\mathcal{K_B}$ (background) and $\mathcal{K'_S}$ (semantically enriched) -- unified into a single descriptor collection $\widetilde{\mathcal{K}}$. Although many matchers (e.g.\ SuperGlue) assume a homogeneous set of descriptors, we concatenate the background and semantically enhanced descriptors for two images $A,B$: 
\[
\widetilde{\mathcal{K}}^A = [\mathcal{K_B}^A, \mathcal{K'_S}^A] \quad \text{and} \quad \widetilde{\mathcal{K}}^B = [\mathcal{K_B}^B, \mathcal{K'_S}^B],
\]
and feed them in a single pass. Let $\mathbf{k}_i^A = (\mathbf{p}_i^A, \mathbf{d}_i^A)$ and $\mathbf{k}_j^B = (\mathbf{p}_j^B, \mathbf{d}_j^B)$ denote the keypoints (with positions and descriptors) in images $A$ and $B$, respectively. Using a (learned or classical) similarity function $\Phi$, we formulate matching as a bipartite assignment problem:

\begin{equation}
\label{eq:match}
    \begin{aligned}
        \hat{\mathcal{M}} \;=\; 
        \underset{\mathcal{M} \subseteq \,\widetilde{\mathcal{K}}^A \times \widetilde{\mathcal{K}}^B}{\arg\max}
        \;\;
        \sum_{(\mathbf{k}_{i}^A, \mathbf{k}_{j}^B) \,\in\, \mathcal{M}}
        \Phi\bigl(\mathbf{d}_i^A, \mathbf{d}_j^B\bigr). \\
        \text{subject to each keypoint being matched at most once.}
    \end{aligned}
\end{equation}


Here, $\Phi(\mathbf{d}_i^A, \mathbf{d}_j^B)$ represents the similarity score between descriptors $\mathbf{d}_i^A$ and $\mathbf{d}_j^B$. We employ SuperGlue for its context-based matching, although the above formulation also holds for other matchers (e.g.\ LightGlue). By consolidating both standard and semantically enriched descriptors into a single inference pass, our approach increases matching reliability while maintaining compatibility with standard visual-SLAM pipelines~\cite{CHENG2022104992}. Further evaluation details are provided in Section~\ref{ssec:mab}.

\section{Experimental Evaluation}
\label{sec:exp}

%
Our KSI approach integrates semantic information into keypoint descriptors to mitigate perceptual aliasing in vineyards. Below, we present our experiments to demonstrate the capabilities of our method. The results support our key claims:
(i) semantic enhancement of keypoint descriptors improves matching accuracy on image regions within masks derived from panoptic segmentation compared to their original form;
(ii) leveraging the improved matching accuracy of semantically enriched descriptors leads to more reliable relative pose estimation during vineyard traversals;
(iii) additional semantic context improves visual localisation accuracy in the same vineyard, outperforming traditional descriptors.
These claims are backed up by our experimental evaluation.



\subsection{Semantically Enhanced Keypoint Matching}
\label{sec:sm}

\begin{table*}[t]
  \caption{Keypoint matching accuracy on semantic instances with and without KSI across different months.}
  \centering
  \begin{tabular}{ccccccccccccc}
    \toprule
    \multirow{2}{*}{\textbf{Matcher}} & \multirow{2}{*}{\textbf{Descriptor}} & \textbf{Avg. Accuracy} & \multicolumn{5}{c}{\textbf{With KSI (\%)}} & \multicolumn{5}{c}{\textbf{Without KSI (\%)}}\\ 
    \cmidrule(lr){4-8} \cmidrule(lr){9-13}
    & & \textbf{Gain \%} & March & April & May & June & September & March & April & May & June & September \\
    \midrule
    \multirow{2}{*}{Superglue} & SuperPoint & 06.70 & \textbf{79.03} & \textbf{82.65} & \textbf{86.48} & \textbf{89.34} & \textbf{94.36} & 75.96 & 76.54 & 80.35 & 80.56 & 84.95 \\ 
    & SIFT & 14.58 & \textbf{71.28} & \textbf{80.49} & \textbf{80.52} & \textbf{86.24} & \textbf{93.80} & 60.39 & 65.30 & 68.03 & 70.80 & 74.93\\ 
    \midrule
    \multirow{2}{*}{LightGlue} & SuperPoint &  04.27 & 74.42 & \textbf{80.43} & \textbf{84.28} & \textbf{87.28} & \textbf{93.38} & \textbf{75.80} & 77.05 & 80.28 & 80.53 & 84.79\\ 
    & SIFT & 10.65 & \textbf{70.87} & \textbf{74.97} & \textbf{81.07} & \textbf{86.06} & \textbf{88.39} & 60.97 & 64.51 & 72.20 & 75.13 & 75.32\\ 
    \bottomrule
  \end{tabular}
  \label{tab:sen}
\end{table*}

The first experiment evaluates the performance of our approach, showing that semantically enhancing the keypoint descriptors improves matching accuracy in semantically significant vineyard image regions compared to the original descriptors. We deployed the KSI framework with SuperGlue and LightGlue to match features from SuperPoint and SIFT descriptors across image pairs sampled from sequences collected in March, April, May, June, and September within the same vineyard rows. We focus on keypoints in semantically significant regions (trunks and buildings) and report the percentage of correctly matched keypoints for each image pair. Because background keypoints lack semantic masks for descriptor enhancement, we do not include them in our analysis. As a baseline, we also performed feature matching without the KSI framework to quantify the semantic matching accuracy gains offered by our method.

\begin{figure}[t]
  \centering
  \includegraphics[width=0.99\linewidth]{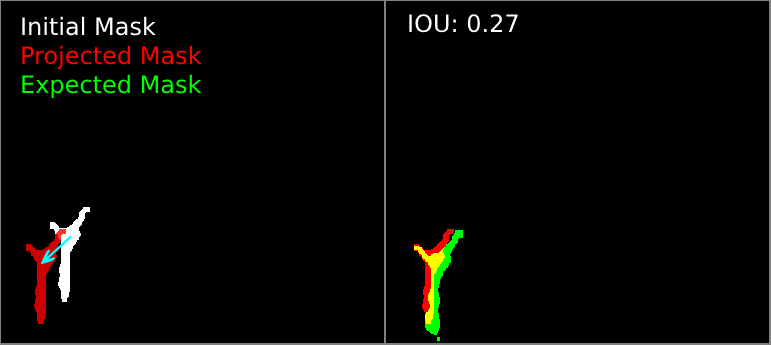}
  \caption{Determination of semantic instance correspondences. The ground truth rotation and translation are applied to the initial mask, producing a projected mask that is then compared with each instance mask in the subsequent frame. The instance with the highest IoU is selected as the corresponding expected mask.}
  \label{fig:iou}
\end{figure}

To verify that a keypoint on a specific trunk or building in one frame is matched to the corresponding semantic instance in the subsequent frame, we established correspondences between semantic instances for each image pair. We use the ground truth rotation and translation to project the semantic instance mask from the first frame onto the next frame in the sequence, to align the mask according to the spatial transformations (as illustrated in Fig.~\ref{fig:iou}). The projected mask was then compared against all semantic instance masks in the subsequent image via the intersection over union (IoU) metric to determine the correct correspondence. A minimum IoU threshold of $0.1$ was applied to filter out false positives, which excluded some distant or smaller object that failed to meet the threshold. While this limits the evaluation to keypoints on closer or larger semantic instances, we still ensure a valid comparison, as the same semantic correspondences are applied to both our method and the baseline.

In Table~\ref{tab:sen}, we compare KSI with baselines that pair SuperPoint or SIFT descriptors and SuperGlue or LightGlue matchers. KSI outperforms the corresponding baselines in semantically significant regions in all four configurations, aside from a single exception in March (SuperPoint + LightGlue), where the difference is under 1\%. The largest improvement appears in September, where SIFT + SuperGlue provides an 18.87\% accuracy gain over its baseline. Overall accuracy tends to increase toward September, likely because heavier foliage reduces distinguishable background keypoints, making semantically enhanced keypoints more valuable. Notably, the highest average accuracy gain (14.58\%) also arises with SIFT + SuperGlue, suggesting that classical descriptors can benefit even more from KSI than learned descriptors like SuperPoint. Nevertheless, the highest absolute accuracy is achieved with by SuperPoint + SuperGlue, and thus we adopt this configuration for all subsequent experiments and ablations.

\subsection{Relative Pose Estimation}

The second experiment evaluates relative pose estimation with KSI-enabled keypoint matching, showing how our method improves pose estimation accuracy for image pairs during vineyard navigation. Accurate relative pose estimation is critical for real-world visual SLAM systems, particularly in agricultural robotics, where repetitive rows structures complicate localisation. By reducing pose errors, KSI-enabled keypoint matching improves trajectory estimation, thus enhancing navigation and task precision in structured environments. 

We compute relative poses between image pairs by estimating the essential matrix via Random Sample Consensus (RANSAC)~\cite{ransac}. The ground truth poses are then used to set the translation scale in the estimated pose. Using these scaled relative pose estimates, we generate a trajectory for each month and compare it against the ground truth trajectory recorded by RTK-GPS. Each test sequence involves the robot traversing two vineyard rows in a loop -- moving along one row, returning along another, and then repeating the forward pass -- yielding a closed-loop path of approximately 153\,m (illustrated in Figure \ref{fig:vin}).  We employ the evo library~\cite{evo} to calculate both the relative pose error (RPE) and the absolute pose error (APE) of the trajectory. RPE and APE results obtained from SuperGlue feature matching with KSI are compared against those produced without KSI.

\begin{table*}[t]
  \centering
  \caption{Comparison of relative pose error (RPE) and absolute pose error (APE) across five months in a 153\,m vineyard loop, using SuperPoint descriptors with SuperGlue matching. The KSI-enabled approach consistently reduces RPE and APE over the baseline, in some cases by more than half, with the largest improvement occurring in September.}
  \begin{tabular}{cccccc}
    \toprule
    \multirow{2}{*}{\textbf{Evaluation Method}} & \multicolumn{5}{c}{\textbf{Month : Mean (Standard Deviation)  [Unit: cm]}} \\
    \cmidrule(lr){2-6}
    & March & April & May & June & September \\
    \midrule
    Relative pose error with KSI & \textbf{17.32(11.49)} & \textbf{8.07(9.02)} & \textbf{5.85(5.71)} & \textbf{5.94(5.59)} & \textbf{3.28(11.20)} \\
    Relative pose error without KSI  & 20.47(5.42) & 12.53(8.39) & 8.97(5.13) & 9.41(5.52) & 11.28(9.92) \\
    \midrule
    Absolute pose error with KSI & \textbf{109.58(91.85)} & \textbf{117.23(76.21)} & \textbf{113.01(63.01)} & \textbf{150.95(78.05)} & \textbf{153.46(98.08)} \\
    Absolute pose error without KSI & 532.82(272.71) & 398.24(290.34) & 274.10(169.69) & 358.52(151.41) & 1013.72(804.68) \\
    \midrule
    RPE Improvement, APE Improvement & 15.39\%, 79.43\% & 35.59\%, 70.56\% & 34.78\%, 58.77\% & 36.88\%, 57.90\% & 70.92\%, 84.86\%\\
    \bottomrule
  \end{tabular}
  \label{tab:rpe}
\end{table*}

The results in Table~\ref{tab:rpe} show that KSI-enabled pose estimation consistently yields lower RPE and APE than the baseline. The largest performance gain appears in September, where the scarcity of distinguishable features makes pose retrieval especially challenging. This finding aligns with our motivation for semantic enrichment, demonstrating its effectiveness in visually limited scenarios.

\subsection{Visual Localisation}

Our third experiment demonstrates how KSI can accommodate cross-seasonal variations in vineyard environments, enabling robust visual localisation despite significant foliage growth. We uniformly sampled 136 images per row from two vineyard rows: row 1, located near the vineyard centre and containing mostly trunks, and row 2, closer to the edge, featuring both trunks and buildings (see Figure~\ref{fig:vin}). Using the hierarchical localisation toolbox~\cite{sarlin2019coarse} for 6-degrees of freedom (DoF) visual localisation, we generate two point clouds from March images: one with SuperPoint+SuperGlue using KSI, and one without KSI. As shown in Table \ref{tab:vis}, our approach delivers a 35.06\% improvement in visual localisation accuracy for row 1, which offers fewer robust features. In contrast, accuracy in row 2 remained similar to the baseline. Hence, KSI significantly enhances cross-seasonal localisation in challenging vineyard scenes while maintaining similar performance under less demanding conditions.

\begin{table}[t]
  \caption{Visual localisation errors relative to March. The SuerPoint descriptors with SuperGlue matcher was used in this comparison. MTE: Median Translation Error in cm, R@$X$: Recall under $X$ meters.}
  \centering
  \resizebox{\columnwidth}{!}{
  \begin{tabular}{cccccc}
    \toprule
    \multirow{2}{*}{\textbf{Metric}} & \multirow{2}{*}{\textbf{Row}} & \multicolumn{4}{c}{\textbf{Month (w/o KSI, with KSI)}}  \\
    \cmidrule(lr){3-6}
    &  & April & May & June & September \\
    \midrule
    \multirow{2}{*}{MTE} & 1 & 23.75, \textbf{11.91} & 21.09, \textbf{13.93} & 24.47, \textbf{16.94} & 513.34, \textbf{381.52} \\
    & 2 & \textbf{06.35}, 06.53 & 07.90, \textbf{07.88} & 06.78, \textbf{06.72} & 13.49, \textbf{11.72} \\
    \midrule
    \multirow{2}{*}{R@0.5} & 1 & 94.85, \textbf{98.53} & \textbf{98.53, 98.53} & \textbf{63.24}, 61.03 & 26.47, \textbf{30.15}\\
    & 2 & 99.27, \textbf{100.00} & \textbf{100.00, 100.00} & \textbf{99.27}, 97.79 & \textbf{91.18, 91.18}\\
    \midrule
    \multirow{2}{*}{R@1} & 1 & \textbf{99.27}, 98.53 & \textbf{98.53, 98.53} & \textbf{64.71}, 62.5 & 28.68, \textbf{30.88} \\
    & 2 & \textbf{100.00, 100.00} & \textbf{100.00, 100.00} & \textbf{99.27}, 98.53 & \textbf{94.85, 94.85} \\
    \midrule
    \multirow{2}{*}{R@5} & 1 & 99.27, \textbf{100,00} & 99.27, \textbf{100.00} & \textbf{72.79}, 70.59 & 32.35, \textbf{35.29} \\
    & 2 & \textbf{100.00, 100.00} & \textbf{100.00, 100.00} & \textbf{99.27, 99.27} & \textbf{96.32, 96.32}\\
    \bottomrule
  \end{tabular}
  }
  \label{tab:vis}
\end{table}


We employed two metrics to assess visual localisation accuracy. First, we computed the Euclidean distance between the reference pose from March and the query pose for each image, then used the median translation error (MTE) over all images in each row. Second, we reported the recall, defined as the percentage of errors below thresholds of 0.5\,m, 1\,m, and 5\,m, while discarding errors exceeding 1km as outliers. The results show that our KSI framework enhances accuracy under seasonal variations caused by foliage growth, especially in scenarios where repetitive structures limit robust feature identification. This is evident from the differing performance levels in row 1 and row 2, which reflects the varying difficulties of each environment.

\subsection{Runtime}

To verify that our model can operate in real-time onboard the robot, we tested it on two devices: (1) a desktop with an 11th Gen Intel i5-11600K 3.90GHz CPU and  NVIDIA GeForce RTX 3070 8GB GPU; (2) and a Jetson AGX Orin with an ARM Cortex-A78 2.2GHz CPU and NVIDIA Ampere 64GB (shared) GPU. 

Table~\ref{tab:tim} presents the runtime results for both systems, including total time,  runtime of the KSI framework and frame rates for the pipeline with SuperPoint and SuperGlue. The desktop achieved the fastest average runtime at 262.8\,ms in September, while the Jetson’s quickest performance was 767.4\,ms in the same month.


\begin{table}[t]
  \caption{Runtime comparison for KSI on Jetson and a desktop computer.}
  \centering
  \begin{tabular}{cll}
    \toprule
    \multirow{2}{*}{\textbf{Component}} & \multicolumn{2}{c}{\textbf{Mean (SD)}} \\
    \cmidrule{2-3}
    & Desktop & Jetson\\
    \midrule
    KSI Framework (ms) & 162.8 (29.7) & 546.9 (74.7)\\
    Total Time (ms) & 304.5 (35.0) & 884.7 (81.1)\\
    Frame Rate (fps)  & 3.3 (0.4) & 1.1 (0.1)\\
    \bottomrule
  \end{tabular}
  \label{tab:tim}
\end{table}

\section{Ablation Study}

We conducted two ablation studies to systematically evaluate and refine the KSI architecture, focusing on: (A) the impact of different semantic integration methods, and (B) the relevance of self- and cross-attention during the matching stage. These studies were intended to optimise the KSI framework and to validate the design choices underpinning our proposed approach.

\subsection{Semantically Enhanced Descriptor Ablation}
\label{ssec:sed}
In this ablation study, we first examine whether adding semantic embeddings to descriptors yields better keypoint matching accuracy in semantically significant regions than simply concatenating them. We then assess the impact of normalisation applied both before and after the addition step, thereby validating and justifying our design choices for implementing normalisation within KSI.

\subsubsection{Addition vs. Concatenation}
Addition and concatenation are two common approaches to incorporate additional information into a feature descriptor. Each approach has its merits for semantic integration: concatenation provides an explicit structure for new semantic information, while addition preserves the original dimensionality and avoids the need for dimensionality reduction when used with a pre-trained matcher. 

When designing feature descriptors for learned matches, one must respect dimensionality constraints, as many matchers are pre-trained on specific descriptor sizes. Ensuring compatibility without retraining is criticql for practical deployment. For example, in our proposed approach using SuperGlue -- trained on $256\times1$ SuperPoint descriptors -- we preserved this dimensionality in the concatenation method by compressing both the semantic masks and original descriptors into $128\times1$ embeddings using autoencoders. We then concatenated these compressed embeddings to form a $256\times1$ descriptor compatible with SuperGlue’s pre-trained network, thereby preserving computational efficiency and avoiding additional matcher retraining.

In contrast, for the addition approach, we compressed the semantic masks into $256\times1$ embeddings and directly added them to the original $256\times1$ SuperPoint descriptors, followed by L2 normalisation of the resulting semantically enriched descriptor. As shown in Table~\ref{tab:cad}, addition outperforms concatenation within the KSI framework. In learned matchers like SuperGlue and LightGlue, addition modifies the descriptor within its original space, aligning more naturally with the matchers' attention mechanisms. Concatenation, on the other hand, partitions the descriptor into separate visual and semantic components, potentially requiring the matcher to adjust to this altered structure. These distinctions can affect how effectively the matchers exploit the enhanced descriptors for feature matching.

\begin{table}[t]
  \caption{Keypoint matching accuracy for different semantic embedding integration approaches in KSI across months. The addition method outperforms concatenation and the baseline without KSI in semantic matching accuracy.}
  \centering
  \begin{tabular}{cccccc}
    \toprule
    \multirow{2}{*}{\textbf{Embed. Method}} & \multicolumn{5}{c}{\textbf{Month (\%)}} \\
    \cmidrule(lr){2-6}
    & March & April & May & June & September \\
    \midrule
    Addition & \textbf{79.03} & \textbf{82.65} & \textbf{86.48} & \textbf{89.34} & \textbf{94.36} \\
    Concat. & 70.49  & 76.59 & 86.41 & 88.04 & 92.82 \\
    w/o KSI & 75.96 & 76.54 & 80.35 & 80.56 & 84.95\\
    \bottomrule
  \end{tabular}
  \label{tab:cad}
\end{table}

\subsubsection{Normalisation of Descriptors}
In the KSI framework, we use an autoencoder to generate semantic embeddings, which are added to the original descriptors. Both the autoencoder bottleneck output and the SuperPoint descriptor can optionally undergo L2 normalisation: semantic embedding normalisation (SN) for the autoencoder embedding, and keypoint descriptor normalisation (KN) for the SuperPoint descriptor. These options yield four possible normalisation configurations, all evaluated in this ablation study to identify the most effective setup for KSI, as shown in Table~\ref{tab:nmz}. To maintain consistency, we applied a mandatory L2 normalisation step after the addition operation so that the final descriptor input to the matcher is always L2 normalised, as indicated in Equation~\ref{eq:sem}. Our analysis in Table~\ref{tab:nmz} shows that the best performance is achieved by not normalising semantic embeddings, and normalising the SuperPoint descriptor. This is because unnormalised semantic embeddings retain their magnitude, thereby shifting the SuperPoint descriptor in the descriptor space. Meanwhile, the post-addition L2 normalisation ensures that the matcher consistently receives a L2-normalised descriptor.

\begin{table}[t]
  \caption{Keypoint matching accuracy on semantic instances for normalisation configurations within KSI. (SN:Semantic Embedding Normalisation, KN:Keypoint Descriptor Normalisation)}
  \centering
  \begin{tabular}{cccccccc}
    \toprule
    \multicolumn{2}{c}{\textbf{Config.}} & \multicolumn{5}{c}{\textbf{Month (\%)}}\\
    \cmidrule(lr){1-2} \cmidrule(lr){3-7} 
    SN & KN & March & April & May & June & September \\
    \midrule
    $\times$ & $\times$ & 78.93 & 82.65 & 86.38 & 89.29 & 94.22 \\
    $\times$ & \checkmark & \textbf{79.03} & \textbf{82.65} & \textbf{86.48} & \textbf{89.34} & \textbf{94.36} \\
    \checkmark & $\times$ & 77.36 & 82.35 & 85.32 & 88.51 & 93.00 \\
    \checkmark & \checkmark & 77.36 & 82.35 & 85.19 & 88.41 & 93.02 \\
    \bottomrule
  \end{tabular}
  \label{tab:nmz}
\end{table}

\subsection{Ablation on Descriptor Matching}
\label{ssec:mab}
We also compare keypoint matching accuracy under a heterogeneous feature matching approach with a classical homogeneous approach. Although SuperGlue was not originally designed for heterogeneous matching, which involves matching descriptors of different types, we adopt it in the KSI framework (see Section \ref{ssec:dm}). Below, we detail why the heterogeneous feature matching outperforms its homogeneous counterpart.

First, we compare the keypoint matching accuracy in semantically significant regions using heterogeneous matching versus a homogeneous matching baseline. In heterogeneous matching, the entire keypoint set $[\mathcal{K_B},\mathcal{K'_S}]$ is processed by a single instance of a matcher. In contrast, the homogeneous baseline employs matches $\mathcal{K_B}$ and $\mathcal{K'_S}$ separately, using two matcher instances, and merges the results to obtain the complete set of matches. Table~\ref{tab:hh} reports the matching accuracy for semantically significant keypoints from $[\mathcal{{K_B}}^A,\mathcal{K'_S}^A]$ in the first image $A$ correspond to $[\mathcal{K_B}^B,\mathcal{K'_S}^B]$ in the second image $B$. The notation $|\mathcal{A}\leftrightarrow\mathcal{B}|$ denotes the percentage of matches between set $\mathcal{A}$ in the image $A$ and set $\mathcal{B}$ in the image $B$. The matching accuracies across semantic and background domains are presented in Table~\ref{tab:mat}.

\begin{table}[t]
  \caption{Keypoint matching accuracy on semantic instances for homogeneous and heterogeneous matcher implementations.}
  \centering
  \begin{tabular}{p{1.5cm} p{0.8cm} p{0.8cm} p{0.8cm} p{0.8cm} p{1.2cm}}
    \toprule
    \multirow{2}{*}{\textbf{Matcher Type}} & \multicolumn{5}{c}{\textbf{Month (\%)}} \\
    \cmidrule(lr){2-6}
    & March & April & May & June & September \\
    \midrule
    Homogeneous & 18.74 & 24.02 & 40.04 & 64.52 & 16.59\\
    Heterogeneous & \textbf{79.03} & \textbf{82.65} & \textbf{86.48} & \textbf{89.34} & \textbf{94.36} \\
    \bottomrule
  \end{tabular}
  \label{tab:hh}
\end{table}

\begin{table}[t]
  \caption{Inter-class and intra-class keypoint matching distribution for heterogeneous matching using a single matcher instance. (S:Semantically significant region, B: Background)}
  \centering
  \begin{tabular}{cccccc}
    \toprule
    \multirow{2}{*}{\textbf{Match Domain}} & \multicolumn{5}{c}{\textbf{Month (\%)}} \\
    \cmidrule(lr){2-6}
    & March & April & May & June & September \\
    \midrule
    $|\mathcal{K'_S}^A \leftrightarrow \mathcal{K'_S}^B|$ & 14.30 & 13.47 & 9.52 & 3.84 & 3.07\\
    $|\mathcal{K'_S}^A \leftrightarrow \mathcal{K_B}^B|$ & 0.14 & 0.21 & 0.11 & 0.04 & 0.04\\
    $|\mathcal{K_B}^A \leftrightarrow \mathcal{K_B}^B|$ & 85.56 & 86.33 & 90.37 & 96.13 & 95.99\\
    $|\mathcal{K_B}^A \leftrightarrow \mathcal{K'_S}^B|$ & 0.0 & 0.0 & 0.0 & 0.0 & 0.0\\
    \bottomrule
  \end{tabular}
  \label{tab:mat}
\end{table}

As shown in Table \ref{tab:hh}, the heterogeneous matching approach outperforms the homogeneous setup that employs separate matcher instances. This substantial overall gain of 53.59\% can be attributed to SuperGlue's GNN context aggregation mechanism, which leverages nearby keypoints to find candidate matches for a given descriptor. By matching the entire keypoint set $[\mathcal{K_B},\mathcal{K'_S}]$ in a one pass -- instead of splitting it into two individual sets -- SuperGlue’s GNN benefits from a broader context, ultimately enhancing matching accuracy.

Table~\ref{tab:mat} further indicates that our heterogeneous matching approach consistently assigns keypoints to their semantic relevance, even though some keypoints are semantically enriched with KSI. Only a small fraction of keypoints in semantically significant regions of the image are matched to background regions ($|\mathcal{K'_S}^A \leftrightarrow \mathcal{K_B}^B|$), while none of the background keypoints match semantically significant regions ($|\mathcal{K_B}^A \leftrightarrow \mathcal{K'_S}^B|$). Table~\ref{tab:mat} also shows how the number of keypoints in semantically significant areas decreases during months with dense foliage. As more foliage covers of the scene, fewer keypoints remain visible in these regions, thus reducing the $|\mathcal{K'_S}^A \leftrightarrow \mathcal{K'_S}^B|$ matches for those months. 

These ablation studies validate our design choices around semantic integration, normalisation, and matching strategy. The finding confirm that incorporating semantic embeddings, applying selective normalisation, and adopting a heterogeneous matching approach significantly enhance keypoint matching accuracy.
\section{Conclusion}
\label{sec:conclusion}

We introduced KSI, a novel method for embedding semantic information into keypoint descriptors to address perceptual aliasing and appearance changes in outdoor environments. 
By enhancing descriptors with semantic context, KSI increases robustness in environments such as vineyards where traditional descriptors often fail due to repetitive structures. Our approach encodes semantic instance masks derived from vineyard-specific objects, producing semantic embeddings that are added to the original descriptors. 
In addition, we presented the SemanticBLT dataset, designed for panoptic segmentation in vineyard environments.
Evaluating our approach across multiple seasons with significant visual and temporal variations, we tested it with various keypoint descriptors and matches. In all descriptor-matcher combinations, our KSI framework achieved an overall 9\% higher matching accuracy in semantically significant regions compared to the respective baselines. Moreover, semantic descriptor enrichment improved both relative pose estimation and visual localisation performance over multiple months, underscoring KSI's effectiveness in challenging agricultural environments.




\bibliographystyle{ieeetr}

\bibliography{glorified,new}

\end{document}